\documentclass[lettersize,journal]{IEEEtran}
\usepackage{amsmath,amsfonts}
\usepackage{algorithmic}
\usepackage{algorithm}
\usepackage{array}
\usepackage[caption=false,font=normalsize,labelfont=sf,textfont=sf]{subfig}
\usepackage{textcomp}
\usepackage{stfloats}
\usepackage{url}
\usepackage{verbatim}
\usepackage{graphicx}
\usepackage{cite}
\usepackage{booktabs}
\usepackage{tabularx}
\usepackage{hyperref}
\usepackage{threeparttable}
\usepackage[switch]{lineno}
\hypersetup{
	colorlinks = true,
	linkcolor = blue,
	citecolor = blue,
}
\usepackage{xcolor}
\usepackage{colortbl}
\usepackage{multirow}
\begin{document}

\title{CV-MOS: A Cross-View Model for Motion Segmentation}

\author{		
	Xiaoyu Tang,~\IEEEmembership{Member,~IEEE,}
	Zeyu Chen,
	Jintao Cheng,
        Xieyuanli Chen,~\IEEEmembership{Member,~IEEE,}
        Jin Wu,
        Bohuan Xue,~\IEEEmembership{Member,~IEEE,}      

        % <-this % stops a space
        
\thanks{This research was supported by the National Natural Science Foundation of China under Grant 62001173, the Project of Special Funds for the Cultivation of Guangdong College Students' Scientific and Technological Innovation (“Climbing Program” Special Funds) under Grant pdjh2022a0131 and pdjh2023b0141.}
        
\thanks{Corresponding author: Xiaoyu Tang. E-mail address: tangxy@scnu.edu.cn.}% <-this % stops a space
\thanks{Xiaoyu Tang, Zeyu Chen and Jintao Cheng is with the School of Electronic and Information Engineering, Faculty of Engineering, South China Normal University, Foshan, Guangdong 528225, China, and also with Xingzhi College, South China Normal University, Guangzhou, Guangdong 510000, China.

Xieyuanli Chen is with the College of Intelligence Science and Technology, National University of Defense Technology, Changsha, China.
xieyuanli.chen@nudt.edu.cn

% Jin Wu is with the Department of Electronic and Computer Engineering, Hong Kong University of Science and Technology, Hong Kong.

Jin Wu are with the Department of Electronic
and Computer Engineering, Hong Kong University of Science and
Technology, Hong Kong, SAR, China.

Bohuan Xue is with the Department of Computer Science and Engineering, Hong Kong University of Science and Technology, Hong Kong, SAR, China
(e-mail: bxueaa@connect.ust.hk)
}}

\markboth{Journal of \LaTeX\ Class Files,~Vol.~14, No.~8, August~2021}%
{Shell \MakeLowercase{\textit{et al.}}: A Sample Article Using IEEEtran.cls for IEEE Journals}

\maketitle

\begin{abstract}
In autonomous driving, accurately distinguishing between static and moving objects is crucial for the autonomous driving system. When performing the motion object segmentation (MOS) task, effectively leveraging motion information from objects becomes a primary challenge in improving the recognition of moving objects. Previous methods either utilized range view (RV) or bird's eye view (BEV) residual maps to capture motion information. Unlike traditional approaches, we propose combining RV and BEV residual maps to exploit a greater potential of motion information jointly. Thus, we introduce CV-MOS, a cross-view model for moving object segmentation. Novelty, we decouple spatial-temporal information by capturing the motion from BEV and RV residual maps and generating semantic features from range images, which are used as moving object guidance for the motion branch. Our direct and unique solution maximizes the use of range images and RV and BEV residual maps, significantly enhancing the performance of LiDAR-based MOS task. Our method achieved leading IoU(\%) scores of 77.5\% and 79.2\% on the validation and test sets of the SemanticKitti dataset. In particular, CV-MOS demonstrates SOTA performance to date on various datasets. The CV-MOS implementation is available at
\href{https://github.com/SCNU-RISLAB/CV-MOS}{https://github.com/SCNU-RISLAB/CV-MOS}
\end{abstract}


\section{Introduction}
\IEEEPARstart For autonomous driving systems, the presence of moving objects creates traces on the map. This significantly undermines the precision of mapping and localization tasks in dynamic environments\cite{song2024ssf}. Accurate identification of these objects greatly aids in accomplishing various downstream tasks such as obstacle avoidance \cite{peters2021inferring}, the SLAM \cite{chen2019suma++} and planning \cite{kummerle2015autonomous}.

Significant progress has been made in the research of traditional semantic segmentation task\cite{kong2023rethinking, wang2023real, li2022holoparser, liu2022heterogeneous, tian20223d, tian2023revised}. However, the semantic segmentation task can only segment stationary objects and cannot handle moving ones. Additionally, due to the irregularity and sparsity of the point cloud, MOS task still faces challenges. This paper follows the mainstream theory of the MOS task, utilizing a projection-based 2D image representation for modeling. It employs projection and residual mapping to segment moving objects within the current frame of the point cloud. Previous projection methods primarily fall into two categories: RV projection\cite{chen2021moving, sun2022efficient, cheng2024mf} or BEV projection \cite{mohapatra2021limoseg, zhou2023motionbev} methods. We will list some examples\cite{sun2022efficient, zhou2023motionbev} to illustrate further.

\begin{figure}
\centering
\includegraphics[width=0.485\textwidth]{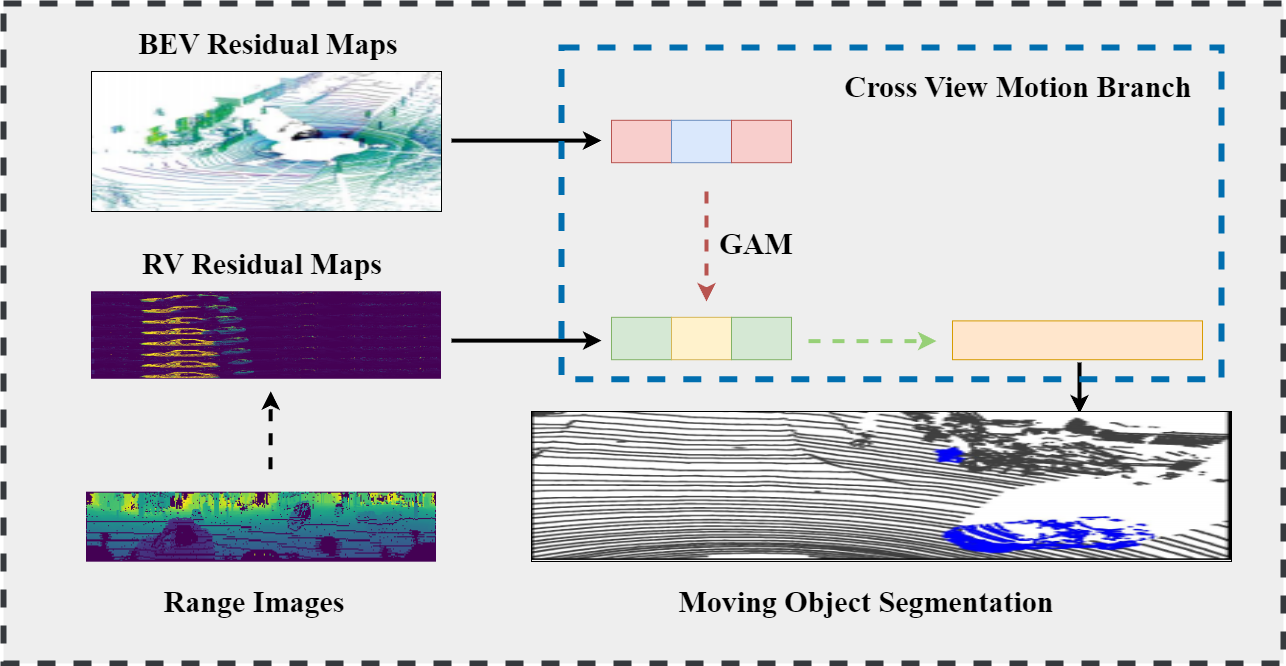}
% [width=0.8\textwidth,height=0.5\textwidth]
\caption{The core idea of the proposed cross-view model. We prioritize leveraging motion information as the primary feature, with semantic information serving as supplementary feature. To achieve this, we propose transitioning from a single-branch motion model to a dual-branch structure. This involves shifting the input source for motion information from a single-view projection to a cross-view motion data, thereby enhancing the representation of motion information.} 
\label{fig: core idea}
\end{figure}

\begin{figure}
\centering
\includegraphics[width=0.485\textwidth]{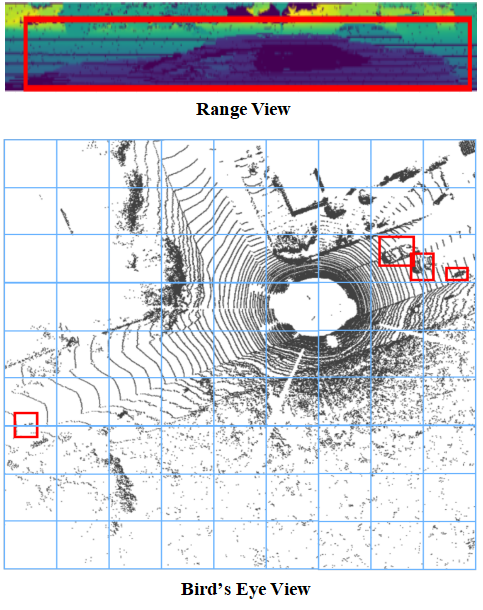}
% [width=0.8\textwidth,height=0.5\textwidth]
\caption{The problems with RV and BEV projection. We can observe that RV projection suffers from issues with occlusion and distorted proportions of objects, leading to imbalance and distortion in object aspect ratios after projection, such as the deformation of a car in a range image. BEV projection does not the above-mentioned issues. However, it introduces quantization error when dividing the space into voxels or pillars, which is unfriendly for distant objects that may only have a few points, such as the small distant object in the bottom left corner of the above picture. } 
\label{fig: problems with rv and bev}
\end{figure}

For instance, MotionSeg3D\cite{sun2022efficient} adopts the dual-branch framework proposed in Salsanext\cite{cortinhal2020salsanext}, encoding both semantic and motion information from RV projection. However, as shown in Fig. \ref{fig: problems with rv and bev}, due to the inevitable occlusion and sensitivity to range associated with RV projection, these methods often suffer from boundary blurring issues during the back-projection process. In addition to utilizing RV projection, some methods employ BEV projection for modeling.  MotionBEV\cite{zhou2023motionbev} projects point cloud into 2D images using BEV projection and utilizes the dual-branch framework proposed in PolarNet\cite{Zhang_2020_CVPR} for modeling. Although BEV projection provides better solutions to occlusion compared to RV projection, as shown in Fig. \ref{fig: problems with rv and bev} and the performance of \cite{zhou2023motionbev} in Tab. \ref{tab: distance} at long distances, it is less effective for distant objects that may only have a few points.

To address the information loss caused by RV and BEV projections, integrating data from multiple view projections can be more effective. Based on intuitive observations, this approach better mitigates the issues associated with single-view projections. Therefore, we propose CV-MOS. The core idea of CV-MOS is to simultaneously utilize the motion information projected by BEV and RV and obtain prediction results in the motion branch (see Fig. \ref{fig: core idea}). We utilize motion features as the primary characteristic, with semantic features as supplementary, and capture moving objects from the motion branch. Building on the traditional dual-branch network structure based on RV projection, we introduce an additional BEV motion branch. This branch takes the residual maps generated by BEV projection as input to encode and capture the motion information from the BEV projection. Moreover, this BEV motion information is integrated with RV motion information. It can enable the model to learn and represent motion information from various views and enhance the model's understanding of object motion information.

Our method shares a similar concept with MF-MOS\cite{cheng2024mf} that captures moving objects from residual maps through a motion branch. Different from \cite{cheng2024mf}, our CV-MOS takes a step further in the representation of motion information by incorporating inputs from two different views. We introduced an additional motion branch to encode the BEV's motion information. The additional motion branch is fused with the motion information encoded by the RV motion branch during the down-sampling process. This effectively mitigates the issue of information loss caused by single-view projection, resulting in superior performance in task completion. Additionally, we introduce a 3D Spatial Channel Attention Module (SCAM), which provides additional attention guidance to the motion branch, thus alleviating potential information loss.

Extensive experiments have demonstrated the superiority of our approach. Leveraging the powerful cross-view motion information representation capability of CV-MOS, we significantly enhance performance on the SemanticKITTI-MOS Dataset\cite{chen2021moving}.

In summary, our contributions are as follows:

$\bullet$ We target the direct capture of the motion information in the MOS task and propose a motion-focused network
with a cross-view structure named CV-MOS: (i) a
primary motion branch to capture the motion feature
from residual maps; (ii) The motion branch has evolved from a traditional single-branch structure to a dual-branch structure with input from different views, allowing for the calculation of object motion information from residual maps of different views.

$\bullet$ We introduce a SCAM to refine the motion branches, alleviate information loss, and improve the model's inference speed.

$\bullet$ This approach achieved the highest ranking on both the validation and test sets of the SemanticKITTI-MOS\cite{chen2021moving} dataset. Additionally, we conducted tests on various benchmark datasets to validate the robustness and superior performance of our method.\\

\section{RELATED WORK}
The earliest approach utilizes prior map-based methods \cite{kim2020remove, xie2020moving, zhuang2023amos} to detect moving objects, which has the advantage of not requiring learning from data.  However, its real-time application is limited due to its reliance on prior maps and pose information obtained from Simultaneous Localization and Mapping (SLAM) systems. Given that deep learning methods learn distribution patterns from historical data, they can detect moving objects without the need for pre-built maps. Therefore, the mainstream approach now is to apply deep learning methods.

In deep learning methods, some point-based approaches \cite{mersch2022receding, sun2020pointmoseg, kreutz2023unsupervised} directly model 3D points to extract spatial-temporal information and improve accuracy. These methods aggregate multi-frame point cloud information, voxelizing the points, and then utilize sparse convolution for feature extraction from voxel blocks. After modeling, voxel blocks are back-projected to points to predict results. However, due to the high dimensionality and sparsity of the 3D point cloud, these methods often suffer from high training costs and slow real-time inference speeds. Therefore, more recent approaches increasingly adopt projection methods, leveraging the lower dimensionality of 2D images to significantly reduce computational costs and spatial requirements. This paper will focus on projection-based methods. These methods can be broadly categorized into two main methods: methods based on RV projection and methods based on BEV projection.

\subsection{Methods Based On RV Projection}
The RV projection method involves projecting 3D points onto a 2D space using spherical mapping. By projecting the point cloud onto range images, the full resolution of LiDAR sensor data is preserved, avoiding spatial loss. Chen et al. \cite{chen2021moving} proposed LMNet, a method that leverages range images and RV residual maps to capture spatial-temporal information and aligns features through direct concatenation. After that, MotionSeg3D\cite{sun2022efficient} was proposed as an improved version, introducing a dual-branch structure. It employs Salsanext\cite{cortinhal2020salsanext} as the encoder for both branches to simultaneously encode semantic and motion information, integrating motion features into semantic features during the encoding process. RVMOS\cite{kim2022rvmos} also illustrated a multi-branch segmentation framework to integrate semantic and motion information. Different from the above methods, MF-MOS\cite{cheng2024mf} focuses more on the feature extraction of motion features, which coincides harmoniously with our viewpoint and has achieved leading performance with additional semantic labels. The above methods all utilize the spatial-temporal information from range images and RV residual maps. However, the RV projection method may encounter boundary blur issues during back-projection and is sensitive to changes in range, which limits the performance of such methods.

\subsection{Methods Based On BEV Projection}
Compared to RV projection, BEV projection offers a more intuitive representation of object motion and spatial relationships within a scene. BEV images exhibit fewer overlaps between objects compared to range images, thus partially addressing boundary blur issues during the back-projection process. Consequently, several methods adopt BEV-based approaches. LiMoSeg\cite{mohapatra2021limoseg} proposes a BEV-based method utilizing disparity and residual computation between two consecutive frames.
After that, MotionBEV \cite{zhou2023motionbev} introduces a dual-branch network that utilized PolarNet \cite{Zhang_2020_CVPR} as the encoder for encoding both motion and semantic information from BEV projections. Additionally, it adopts a window-based approach for generating BEV images, significantly enhancing the performance of MOS task that rely on BEV projection methods. 
However, unlike range images, BEV images are highly sensitive to height variations and prone to compounded errors, resulting in suboptimal performance in detecting smaller objects such as pedestrians and cyclists. Consequently, these limitations also restrict the development of these methods to some extent.

In other domains of point cloud research, methods employing cross-view approaches such as \cite{qiu2022gfnet, ma2023cvtnet} have been utilized for various downstream tasks. For instance, in semantic segmentation, \cite{ma2023cvtnet} uses transformers as encoders for two different views, followed by feature fusion using the proposed Word-Aligned module. However, similar methods have not yet been explored for the MOS task. Effective utilization of motion information is crucial for MOS, and cross-view frameworks in other fields are focused on semantic information rather than motion information. Thus, directly applying cross-view techniques from other domains may not be suitable.

This paper reconsiders the shortcomings of RV and BEV projections and designs a cross-view network model that integrates motion information from both BEV and RV residual maps, capturing moving objects from the motion branch.

\section{OUR APPROACH}
In the subsequent sections, we provide a detailed overview of our CV-MOS. We begin with an extensive overview of the RV and BEV projection techniques. This will cover the methods used for projecting LiDAR point cloud data onto range images, as well as onto RV and BEV residual maps. Following this, we will discuss the network architecture of the proposed CV-MOS in depth.

\begin{figure*}
\centering
\includegraphics[width=1\textwidth]{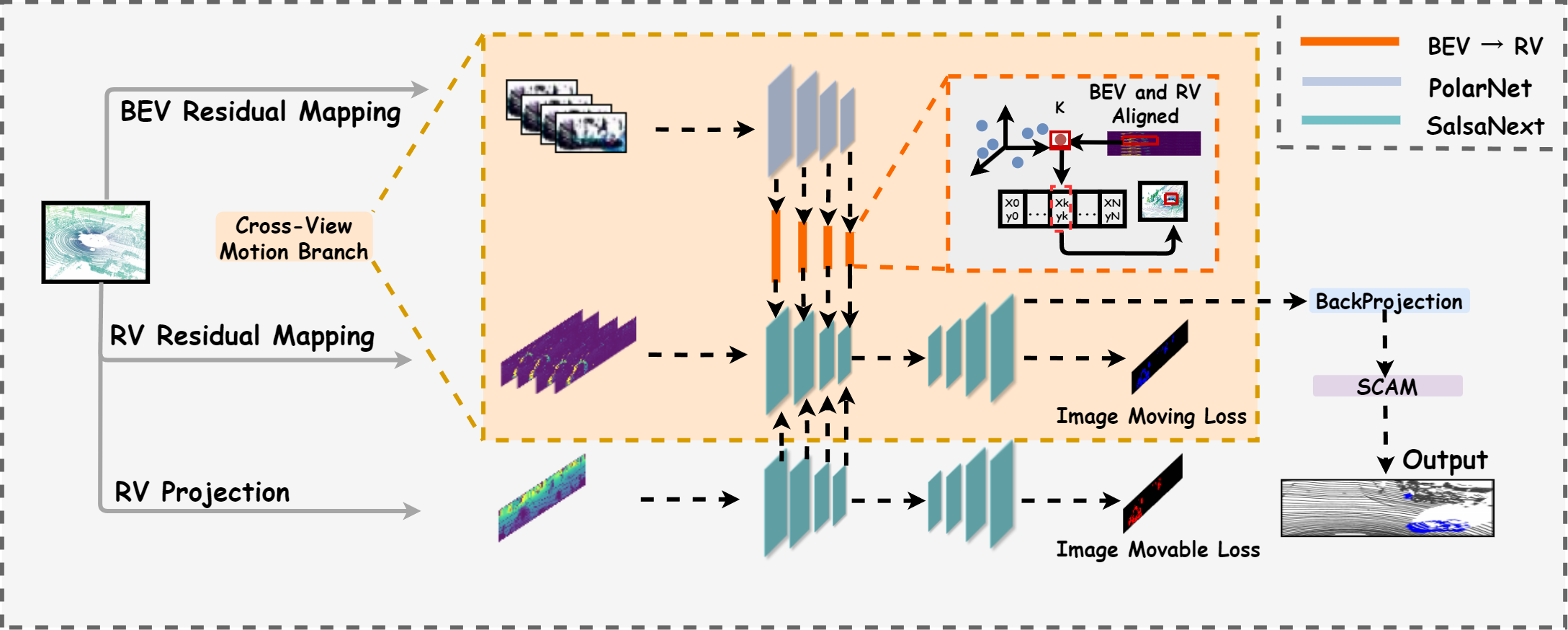}
% [width=0.8\textwidth,height=0.5\textwidth]
\caption{CV-MOS is a cross-view network model with three inputs and two outputs. Range images along with RV and BEV residual maps are respectively fed into their designated branches. Both RV and BEV residual maps are jointly provided to the motion branch to facilitate the fusion of motion features. Initially, the RV motion branch outputs the first-stage prediction results. To achieve more refined segmentation, the final layer features of the RV motion branch are input into the SCAM, which then produces the final point cloud segmentation results.} 
\label{Fig-backbone}
\end{figure*}

\subsection{RV and BEV Projection}
Given a LiDAR point cloud $P \in R^{N\times4}$
, we then have the format of each point as (x, y, z, e), where (x, y, z) is the cartesian coordinate of the point relative to the LiDAR sensor, and e indicates the intensity of returning the laser beam.

For projection-based methodologies, The point cloud must first be transformed into an image $I \in R^{HW\times C}$
to utilize deep neural networks that are predominantly developed for 2D visual recognition. Here, H and W represent the spatial dimensions of the projected images, and C denotes the number of channels. We detail the process for generating range images, RV and BEV residual maps as follows:

\textbf{Range Images.} Following the setup in the previous work \cite{chen2021moving}, for each point (x, y, z) in the 3D point cloud, we employ spherical coordinate projection to project it into a range image, defined by the following formula:
\begin{equation}
\left(\begin{array}{c}u \\ v\end{array}\right) = 
\left(\begin{array}{c}\frac{1}{2}\left[1 - \arctan(y,x)\pi^{-1}\right] w \\ 
\left[1 -(\arcsin(zr^{-1}) + f_{\textup{up}})f^{-1}\right]h\end{array}\right)
\end{equation}
In the given context, (u, v) represents the coordinates of the image, and (h, w) denotes the desired height and width of the range image. 
Here, $f = f_{\textup{up}} + f_{\textup{down}}$ signifies the vertical field of view of the sensor, and $r = \sqrt{x^2 + y^2 + z^2}$ denotes the range of each point. After that, we can use (u, v) to index
the 3D point cloud and integrate its coordinates (x, y, z), the range r, and the intensity e as the five channels of the range image. The range image can be represented by Eq (\ref{equ-2}).

\begin{equation}
\begin{aligned}
I^{k} _{RV} (u_{i} ,v_{i})=[X_{i} ^{k} ,Y_{i} ^{k},Z_{i} ^{k},r_{i} ^{k},e_{i} ^{k}],  \\
k\in\left \{1,2,3,\cdots,K \right \},  \\
i\in\left \{1,2,3,\cdots, N\right \}.
\label{equ-2}
\end{aligned}
\end{equation}

\textbf{RV Residual Maps.} We generate RV residual maps to leverage the spatial-temporal information from continuous LiDAR scans. After obtaining the range images of different frames of the point cloud, we obtain the past k frames residual map $I_{res}$ by calculating the pixel-level variance between frames as follows:

\begin{equation}
\begin{aligned}
I^{k} _{res} (u_{i},v_{i})=\left\{\begin{array}{c}
     \left | \frac{I^{k}_{RV}(u_{i},v_{i})-I^{0}_{RV}(u_{i},v_{i})}{I^{0}_{RV}(u_{i},v_{i})}  \right |,i\in \text{valid pixels}  \\
     0, \text{otherwise} 
\end{array}\right.
\label{equ-3}
\end{aligned}
\end{equation}

Furthermore, we incorporate the data augmentation technique proposed by \cite{cheng2024mf}. Instead of using the single-frame stride, we utilize multi-frame strides to generate RV residual maps. Additionally, we select a 
$\triangle$t at each training iteration instead of depending on a predetermined distribution probability.

\textbf{BEV Residual Maps.} The first step in generating BEV residual maps is to produce BEV  images. Initially, for points (x, y, z) represented in Cartesian coordinates in space, we transform them into polar coordinates.
\begin{equation}
\left(\begin{array}{c}\rho_{j} \\ \theta_{j} \\ z_{j}\end{array}\right) =\left(\begin{array}{c}\sqrt{x_{j}^{2} + y_{j}^{2}} \\ \arctan(y_{j}, x_{j})  \\ z_{j}\end{array}\right)
\end{equation}
Each point $p_{j}$ is assigned to the corresponding grid in the BEV image based on its polar coordinates:
\begin{equation}
\begin{array}{c}
S_{(x, y), i} = \{p_{j} | p_{j} \in S_{i},\\ 
\frac{\rho_{max} - \rho_{min}}{w} \cdot (x - 1) \leq \rho_{j} < \frac{\rho_{max} - \rho_{min}}{w} \cdot x\\ 
\frac{\theta_{max} - \theta_{min}}{h} \cdot (y - 1) \leq \theta_{j} < \frac{\rho_{max} - \rho_{min}}{h} \cdot y\}
\end{array}   
\end{equation}

$S_{(x, y), i}$ represents all the points in the BEV image, where (h, w) denotes the height and width of the BEV image. $(\theta_{min}, \theta_{max})$ and $(\rho_{min}, \rho_{max})$ denote the angular and radial ranges. Points falling outside these specified ranges will be excluded from the point cloud. Subsequently, BEV residual maps will be created based on the generated BEV images.

We follow the setup of MotionBEV\cite{zhou2023motionbev}, using a time window to generate the BEV residual maps, thereby avoiding direct frame-by-frame comparison. We maintain two adjacent time windows, Q1 and Q2, each with a length of $N2 = N/2$. We generate BEV residual maps by examining the height differences of corresponding grid cells in Q1 and Q2. First, we transform the coordinate systems of Q1 and Q2 into the current frame's coordinate system for self-compensation. Then, for each pixel point $I_{(x, y)}$ and BEV image pixel point $Q_{(x, y)}$, the expression of pixel point  
$I_{(x, y)}$ is as follows:
\begin{equation}
\begin{array}{c}I_{(x, y), i} = Max\{Z_{(x, y), i}\} - Min\{Z_{(x, y), i}\}, \\
Z_{(x, y), i} = \{z_{j} \in p_{j} | p_{j} \in Q_{(x, y), i},  z_{min} < z_{j} < z_{max}\}\end{array}
\end{equation}

As for the range of z, we set the range of z within (-4, 2) because this encompasses the most common positions of moving objects of interest, such as pedestrians and vehicles. We subtract the projected BEV images $I_{1}$ and $I_{2}$ to obtain the BEV residual maps:
\begin{equation}
\resizebox{0.45\textwidth}{!}{$
\begin{aligned}
D_{(x, y),i}^0,D_{(x, y),i-1}^1,...,D_{(x, y),i-N_2+1}^{N_2-1}=I_{(x, y),1}-I_{(x, y),2},\\
D_{(x, y),i-N_2}^{N_2+1},D_{(x, y),i-N_2-1}^{N_2+1},...,D_{(x, y),i-N+1}^{N-1}=I_{(x, y),2}-I_{(x, y),1}
\end{aligned}
$}
\end{equation}

In $D_{(x, y), i}^k$, the k represents the $k^{th}$ channel, indicating the motion information when the $i^{th}$ frame serves as the current frame for the preceding k frames. For example, if a residual map is generated, where a pixel (x, y) has $[k_1, k_2, ...k_i]$ channels, then $k_i$ denotes its motion information considering the preceding $i$ frames as the current frame. We shift both temporal windows to the next position as each new frame arrives. Each frame stays within the temporal window for $N$ cycles, thus ensuring that the BEV motion features retain a sequential characteristic with a length of $N$.

\subsection{Network Structure}
We propose a novel cross-view network architecture illustrated in Fig. \ref{Fig-backbone}. The input of CV-MOS consists of range images, RV and BEV residual maps. To explore motion information
for MOS, we modified the traditional dual-branch network structure into a tri-branch one, with the additional motion branch dedicated to encoding the motion information of BEV projection. Specifically, the contributions of CV-MOS proposed in this paper are as follows:

\subsubsection{The Cross-View Motion Branch Structure}
We have designed a cross-view motion branch structure to replace the traditional single-view motion branch structure. As shown in Fig. \ref{Fig-backbone}, the motion information of BEV and RV is input into different encoders for encoding. The encoded BEV motion features are input into the BEV motion branch and fused with the RV motion features to obtain the final motion features. Compared with traditional single-view motion branch encoding, the cross-view motion branch structure can more effectively utilize motion information and reduce information loss caused by projection.

\subsubsection{The Cross-View Motion Features Encoding}
Effectively utilizing motion information is crucial for the MOS task. The encoding of this motion information is primarily divided into two key parts: encoding motion information derived from RV  projection and encoding motion information from BEV projection. Here's how the encoding of motion information from RV projection is carried out:
\begin{equation}
\left\{\begin{matrix}F_{s}=\text{sigmoid}\left(\text{Conv}_{1\times1}\left(F_{\text{rv-semantic}} \right ) \right ) \cdot F _\text{rv-motion},
\\F_{f}=\mathrm{softmax}\left(\mathrm{Conv}_{1\times1}\left(Pool\left(\boldsymbol{F}_{s}\right)\right)\right) \cdot C,
 \\F_{rv}=F_{c}+F_{\mathrm{res}}
\end{matrix}\right.
\label{dual}
\end{equation}
Here, \textbf{$\cdot$} represents the dot product of matrices. The output result $F_{rv}$ of Eq (\ref{dual}) is the fusion of the RV semantic feature map and the RV residual feature map. 

The encoding of motion features from BEV projection is as follows:
\begin{equation}
F_{bev} = \text{Conv}_{3\times3}(\text{Conv}_{3\times3}(\text{MaxPool2d}(F_{\text{bev-motion}}))
\end{equation}

$F_{bev}$ represents the motion features extracted using Polarnet\cite{Zhang_2020_CVPR}. After obtaining the motion features from two different views, $F_{rv}$ and $F_{bev}$, cross-view feature fusion can be performed.

\begin{figure}
\centering
\includegraphics[width=0.485\textwidth, trim=-20 0 0 0, clip]{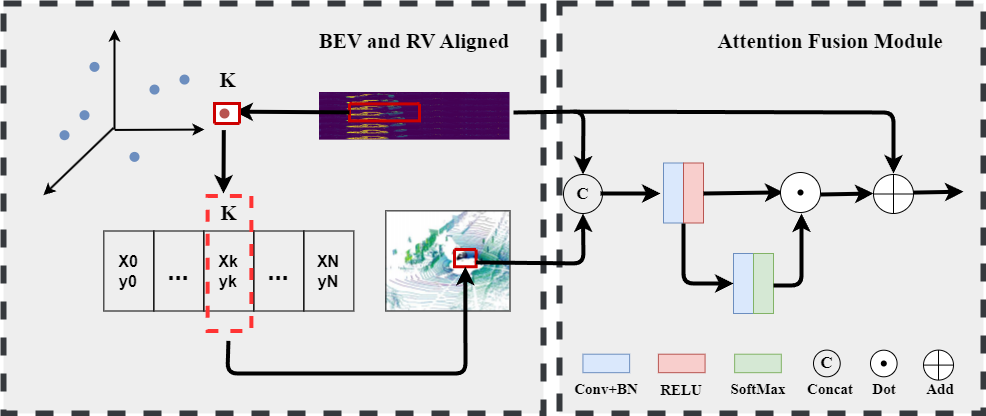}
% [width=0.8\textwidth,height=0.5\textwidth]
\caption{The range image first locates the position index of its corresponding 3D point through indexing, then extracts the points from the corresponding BEV residual map based on the position index, and merges the RV and BEV residual feature map into the corresponding positions in the range image. Fusion is then performed in the following attention fusion network. } 
\label{fig: cross view}
\end{figure}

The overview for integrating various views is illustrated in Fig. \ref{fig: cross view}. The key to this process is how to obtain the transformation matrix ${T}_{B\to R}$ (BEV to RV). 
We employ the geometric calibration method outlined in GFNET \cite{qiu2022gfnet} for alignment. The fundamental concept revolves around establishing the geometric transformation matrix between two views. To obtain $T_{B\to R}$ transformation matrix, We begin by utilizing back-projection to generate an index matrix for each point in the range image, that corresponds to the original points in the point cloud:
\begin{equation}
\mathbf{T}_{R\to P}=\begin{bmatrix}pointidx_{i}&\cdots&pointidx_{j}\\\vdots&\vdots&\vdots\\pointidx_{k}&\cdots&pointidx_{n}\end{bmatrix}
\end{equation}

The shape of the ${T}_{R\to P}$ (from range image to point) matrix is $(H, W)$, where $(H, W)$ represents the height and width range of the range image, and the elements of ${T}_{R\to P}$ matrix, denoted as $pointidx_{k}$, signify the index of the original point cloud corresponding to each point on the range image. If multiple points are projected onto the same pixel, only the point with the smaller range is preserved. Additionally, if any point does not project to a pixel, it's replaced by -1. 

Next, we utilize BEV projection to obtain the ${T}_{P\rightarrow B}$ (point to BEV) matrix:
\begin{equation}
\begin{aligned}
\mathbf{T}_{P\rightarrow B}& =\begin{bmatrix}x_0&\cdots&x_{N-1}\\y_0&\cdots&y_{N-1}\end{bmatrix}^T  
\end{aligned}    
\end{equation}

The shape of ${T}_{P\rightarrow B}$ is (N, 2), where $N$ denotes the total number of points in the 3D point cloud, and $(x_{i}, y_{i})$ denotes the coordinates of the projection of each point in the 3D point cloud onto the BEV image.

After acquiring the ${T}_{R\to P}$ and ${T}_{P\rightarrow B}$ matrices, we utilize the values of each point in the ${T}_{R\to P}$ matrix as indices to access ${T}_{P\rightarrow B}$. Subsequently, we substitute the original values of ${T}_{R\to P}$ with the outcomes obtained from this indexing process. This leads to the generation of the final $T_{B\to R}$(BEV to RV) matrix:

\begin{equation}
\begin{aligned}
\mathbf{T}_{B\rightarrow R}& =\begin{bmatrix}px_{0}&\cdots&px_{H_{r-1}}\\py_{0}&\cdots&py_{W_{r-1}}\end{bmatrix}^T  
\end{aligned}    
\end{equation}

The shape of ${T}_{B\to R}$ matrix is (H, W, 2), where (H, W) represents the width and height of the matrix, which aligning with the range image. The elements in the last dimension, "2," are described as $(px_{i}, py_{i})$, signifying the coordinates in the BEV image corresponding to each point in the range image.
After obtaining the ${T}_{B\to R}$ matrix, we can, for each position $(i, j)$ in range image, obtain its corresponding coordinates $(px_{i}, py_{j})$ in the BEV image through ${T}_{B\to R}[i, j]$. Due to the one-to-one correspondence between residual mapping and points in the image, we can use the matrix ${T}_{B\to R}$ to integrate the features from the BEV residual feature map into the RV residual feature map. This process results in the transformed feature map $M_{B\to R}$, which retains the same shape as the RV residual feature map.  

After transforming the BEV residual feature map into the RV residual feature map using the ${T}_{B\to R}$ matrix, it can enter the network for feature fusion. For the choice of fusion module, we opt to build upon the GFNET\cite{qiu2022gfnet} fusion module by adding a 1$\times$1 convolutional structure, enabling a more fine-grained fusion of features from two different views. The feature fusion process between different views is shown by the following Algorithm \ref{algo: diff views fusion}.
\begin{algorithm}[H]
% \scalebox{0.8}{
Input: $\left\{\begin{array}{@{}l@{}}
\text{BEV feature maps } M_{b}: [Batch, H_{b}, W_{b}, C_{b}], \\
\text{RV feature maps } M_{r}: [Batch, H_{r}, W_{r}, C_{r}], \\
\phantom{\text{RV feature maps }} {T}_{B\to R}: [Batch, H_{r}, W_{r}, 2]
\end{array}\right\}$

Output: $
\text{Fusion feature maps } M_{r}: [Batch, H_{r}, W_{r}, C_{r}], \\
$

Step 1: Geometric Alignment\\
${T}_{(B\to R)} = {T}_{B\to R}.\text{permute(0, 3, 1, 2)}, \\
{T}_{(B\to R)} = \text{F.interpolate}({T}_{(B\to R)}, (H_{r}, W_{r})),\\
{T}_{(B\to R)} = {T}_{B\to R}.\text{permute(0,2,3,1)},\\
M_{B \to R} =  \text{F.grid\_sample}(M_{b}, {T}_{(B\to R)})
$\\

Step 2:  Attention Fusion Module: \\
$
M_{fusion}$=Concat$((M_{r}, M_{B \to R})),\\
M_{fusion}$=$Conv_{3\times3}(Conv_{1\times1}(M_{fusion})),\\
M_{r}=M_{fusion} \cdot Conv_{1\times1}(M_{fusion}) + M_{r}
$
\caption{Geometric Attention Module}
% }
\label{algo: diff views fusion}
\end{algorithm}

\subsubsection{Spatial and Channel Attention Module}
\begin{figure}[H]
    \centering
    \includegraphics[width=0.485\textwidth]{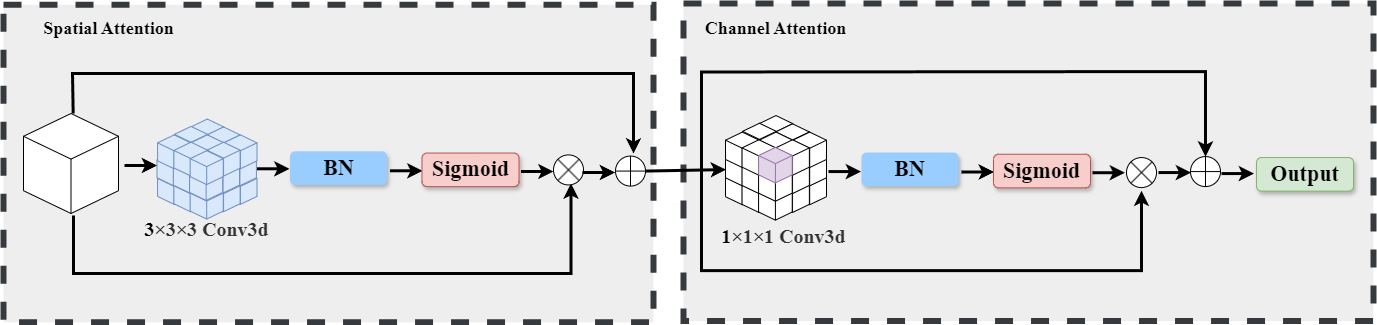}
    \caption{The SCAM, through spatial and channel attention mechanisms, guides voxel blocks to retain crucial information while filtering out interfering information.}
    \label{fig: SCA}
\end{figure}

To compensate for the information loss caused by the dimension reduction process from point cloud to range image, based on the MotionSeg3D\cite{sun2022efficient} second-stage voxel model, we propose the SCAM to refine the segmentation results of the CV-MOS. First, the final layer features of the RV motion branch are back-projected to 3D space, resulting in a feature vector $F_{p}$ with shape $(N, C)$, $N$ denotes the number of points, and $C$ denotes the feature dimension. After voxelizing $F_{p}$, it is input into our SCAM for further processing. As shown in Fig. \ref{fig: SCA}, we use $SparseConv_{3\times3}$ and $SparseConv_{1\times1}$ for spatial and channel feature extraction respectively, followed by a sigmoid activation function to obtain two attention score matrices $F_{s}$ and $F_{c}$. These two matrices represent the spatial attention score matrix and the channel attention score matrix, respectively. These attention score matrices are then dot product applied with the original input to reassemble the original features, retaining key information while filtering out redundant information. 

Finally, we apply a de-voxelization operation to the processed point cloud and fuse it with the point cloud data after MLP feature extraction. Afterward, a classifier is employed to output the refined per-point segmentation results. 

\subsection{Loss Function}
Since the BEV motion branch only involves down-sampling, it does not include the calculation of loss. The loss function applies to both the RV motion branch and the RV semantic branch.

The loss function follows the settings of \cite{cheng2024mf}, including losses from the RV motion and the RV semantic branch. The total loss can be defined as:
\begin{equation}
\mathcal{L}_{\mathrm{Total}}=\mathcal{L}_{\mathrm{RV-Semantic}}+\mathcal{L}_{\mathrm{RV-Motion}}
\end{equation}

Where $\mathcal{L}_{\mathrm{RV-Semantic}}$ and $\mathcal{L}_{\mathrm{RV-Motion}}$ represent the loss of the RV semantic branch and RV motion branch. Both consist of cross-entropy loss $\mathcal{L}_{\mathrm{wce}}$ and Lovasz-Softmax loss $\mathcal{L}_{\mathrm{ls}}$, the loss for each branch is:

\begin{equation}
\mathcal{L} = {L}_{\mathrm{wce}} + \mathcal{L}_{\mathrm{ls}}
\end{equation}

\section{EXPERIMENTS}
We conducted extensive experiments to demonstrate the effectiveness of the CV-MOS. We selected two datasets, SemanticKITTI-MOS\cite{chen2021moving} and Apollo \cite{lu2019l3} dataset, to evaluate the model's generalization ability. Moreover, we have rigorously designed ablation experiments to verify the effectiveness of our method. In the following sections, we will first introduce the experimental setups of the MOS task in detail and then report our experimental results.

\subsection{Experiment Setups}
\textbf{Datasets.} Since the SemanticKITTI-MOS\cite{chen2021moving} dataset is currently the largest and most widely used dataset, it serves as the primary dataset for evaluating the model's performance. We followed the settings of previous work \cite{chen2021moving} for the training, validation, and testing of CV-MOS. Additionally, we also perform validation experiments on the Apollo\cite{lu2019l3} dataset, accompanied along with quantitative analysis following the standard experiment setting in \cite{chen2022automatic}

\textbf{Implementation details.} Our code is implemented in PyTorch. The model training is divided into two stages. In the first stage, we use 2D labels for supervised training. 2D labels include RV moving and semantic labels. In the second stage, we freeze the parameters of the 2D segmentation network and use 3D sparse convolution with our proposed SCAM for supervised training.

During the first stage of training, we employ distributed data-parallel training using 4 NVIDIA RTX 4090 GPUs to implement our method. A total of 150 epochs of training are conducted with an initial learning rate of 0.01 and a decay factor of 0.99 per epoch. The batch size per GPU is set to 4. We use an SGD optimizer with a momentum of 0.9 during training. In the second stage, training is performed using 1 NVIDIA RTX 4090 GPU with an initial learning rate of 0.001. A total of 5 epochs of training are conducted using a batch size of 1, on 1 GPU, and other hyperparameters remained consistent with the first stage.

% \textbf{Implementation details.} 
% Our code is implemented in PyTorch. The experiments are conducted on 4 NVIDIA 4090 GPUs. The batch size is set to 4 for each GPU. We will keep the other hyperparameter settings of the training consistent with \cite{sun2022efficient}.

\subsection{Comparison with State-of-the-Art Methods}
We first report the validation and test results on the SemanticKITTI-MOS\cite{chen2021moving} dataset in Tab.\ref{compare-sota}. We achieved SOTA performance on both the validation and test sets. Particularly, our performance on the test set, 79.2\%, significantly outperforms other methods. Compared to the last SOTA\cite{cheng2024mf}, CV-MOS achieves a 2.5\% higher IoU on the test set, demonstrating the robustness of our model.

\begin{table}
\centering
\setlength{\tabcolsep}{4pt}
{\fontsize{11}{14}\selectfont
\caption{COMPARISONS RESULT ON SEMANTICKITTI-MOS DATASET.}
\begin{tabular}{lccc}
\toprule %添加表格头部粗线
Methods   & Year     & Validation(\%) & Test(\%) \\ 
\midrule

LMNet\cite{chen2021moving}    &2021        & 63.8           & 60.5     \\
LiMoSeg\cite{mohapatra2021limoseg} &2021  & 52.6           & -        \\
Cylinder3D    &2021   & 66.3           & 61.2     \\
RVMOS\cite{kim2022rvmos}   &2022          & 71.2           & 74.7     \\
4DMOS\cite{mersch2022receding}   &2022          & 77.2           & 65.2     \\
MotionSeg3D\cite{sun2022efficient}  &2022    & 71.4           & 70.2     \\
InsMOS\cite{Wang2023InsMOSIM}   &2023        & 73.2           & 75.6     \\  
SSF-MOS\cite{song2024ssf}   &2024            & 70.1           & -
\\
MotionBEV\cite{zhou2023motionbev}    &2023        & 76.5           & 75.8     \\ 
3D-SeqMOS   &2024     & -              & 74.9 \\
MF-MOS\cite{cheng2024mf}    &2024        & 76.1           & 76.7     \\ 
\midrule
Ours        &\_            & \textbf{77.5}  & \textbf{79.2}     \\ 
\bottomrule %添加表格底部粗线
\label{compare-sota}
\end{tabular}
}
\end{table}

To delve deeper into the advantages conferred by our approach, we undertook an exhaustive comparison of segmentation performance among existing methodologies on the SemanticKITTI-MOS validation set across varying distances. As shown in Tab~\ref{tab: distance}, our CV-MOS demonstrates superior performance over other projection methods\cite{sun2022efficient, chen2021moving, zhou2023motionbev, cheng2024mf} at medium to long distances. Within the 0-20m distances, CV-MOS has shown improvement compared to RV projection\cite{sun2022efficient, chen2021moving, cheng2024mf} methods but slightly lags behind the method in \cite{zhou2023motionbev}. This is due to our model’s primary reliance on RV projection for motion and semantic information, with BEV projection as a complementary source. While BEV-based methods are slightly superior to RV-based methods at close distances, their performance diminishes significantly at longer distances. Therefore, we have opted to prioritize RV information as the primary data source for our model.

\begin{figure}
% \centering
\includegraphics[width=0.485\textwidth]{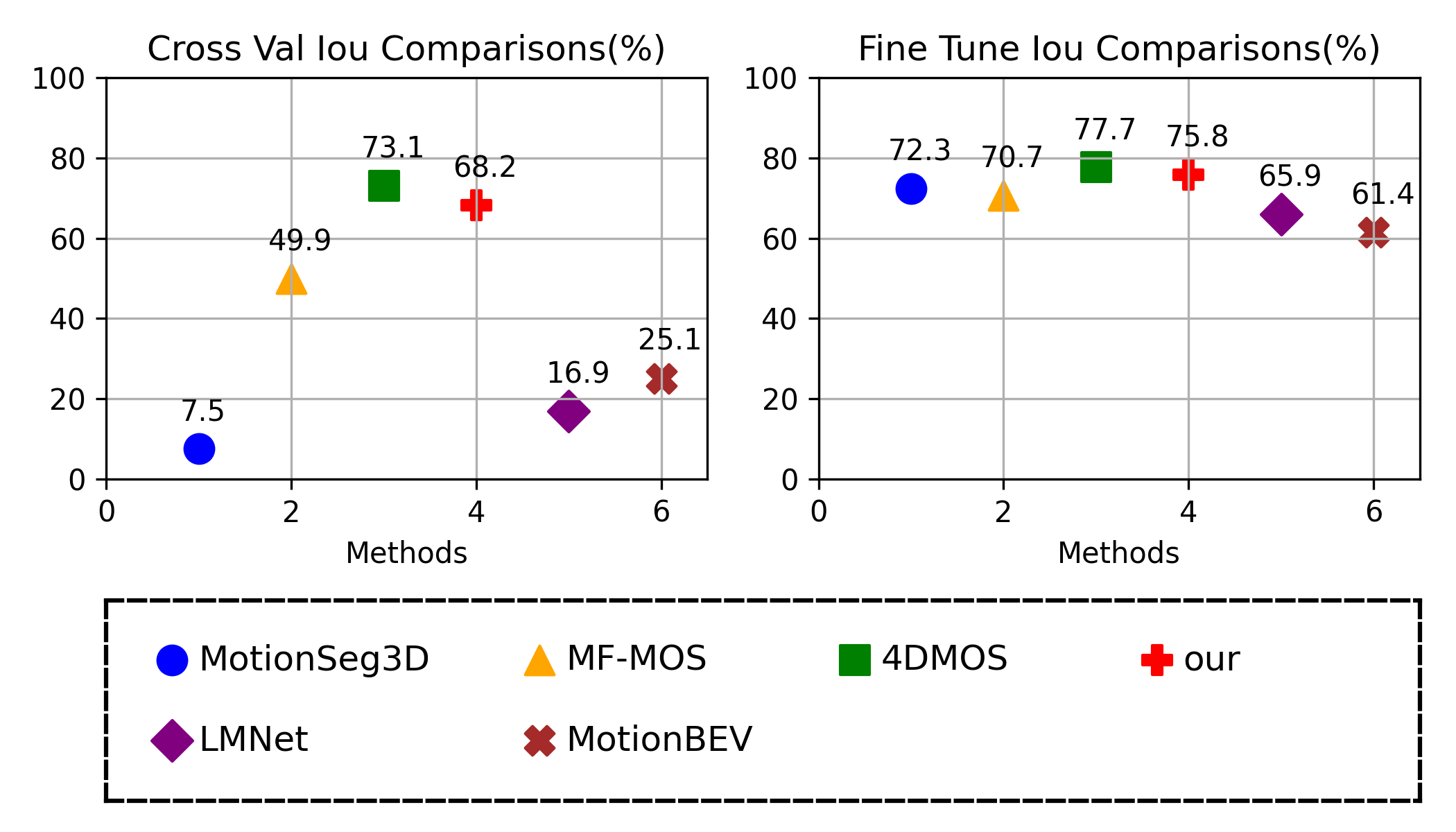}
\hspace{1cm}
% [width=0.8\textwidth,height=0.5\textwidth]
\caption{Apollo dataset performance comparisons.} 
\label{fig: apollo result}
\vspace{0.1cm}
\end{figure}

\begin{figure*}
\centering
\includegraphics[width=1\textwidth]{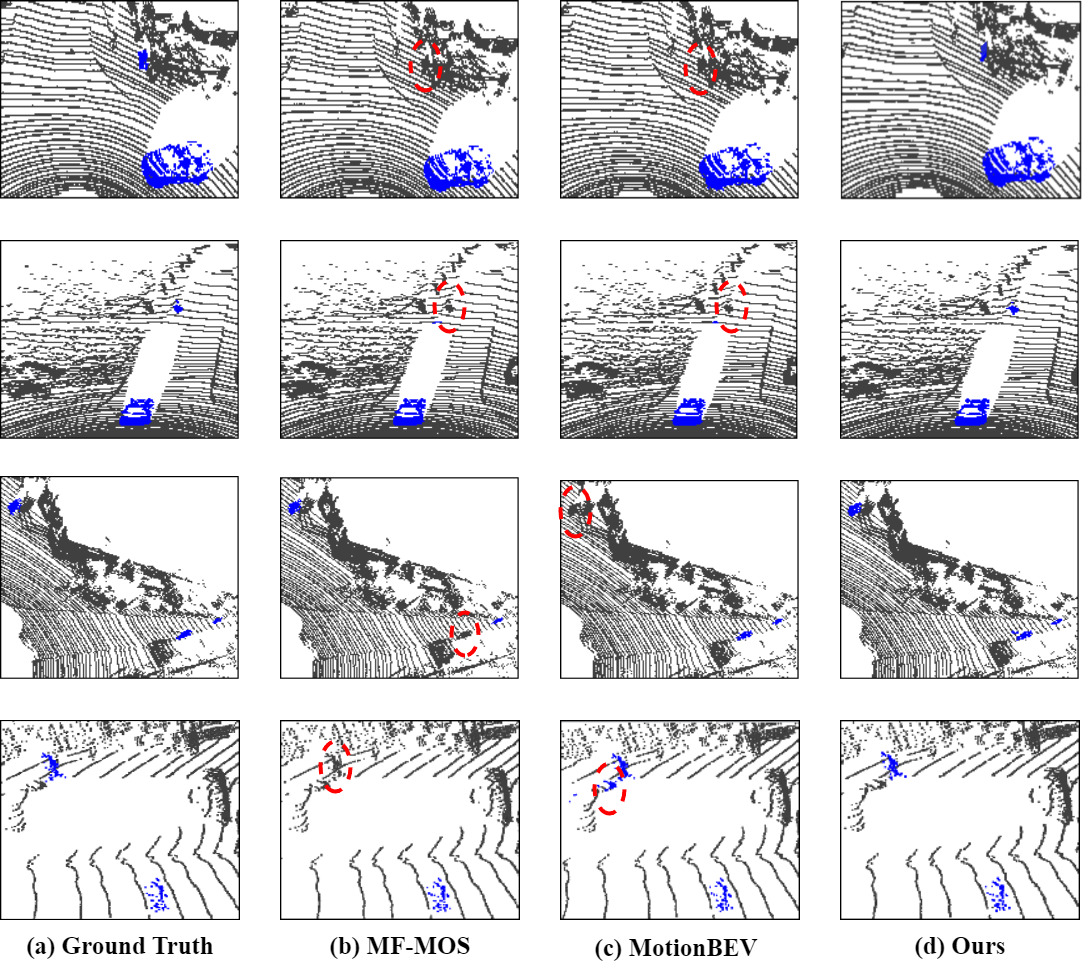}
% [width=0.8\textwidth,height=0.5\textwidth]
\caption{Qualitative results of different methods for LiDAR-MOS on the validation set of the SemanticKITTI-MOS dataset. The red circles highlight incorrect predictions and blurred boundaries.} 
\label{Fig-visualize}
\end{figure*}

\begin{figure}
\centering
\includegraphics[width=0.485\textwidth]{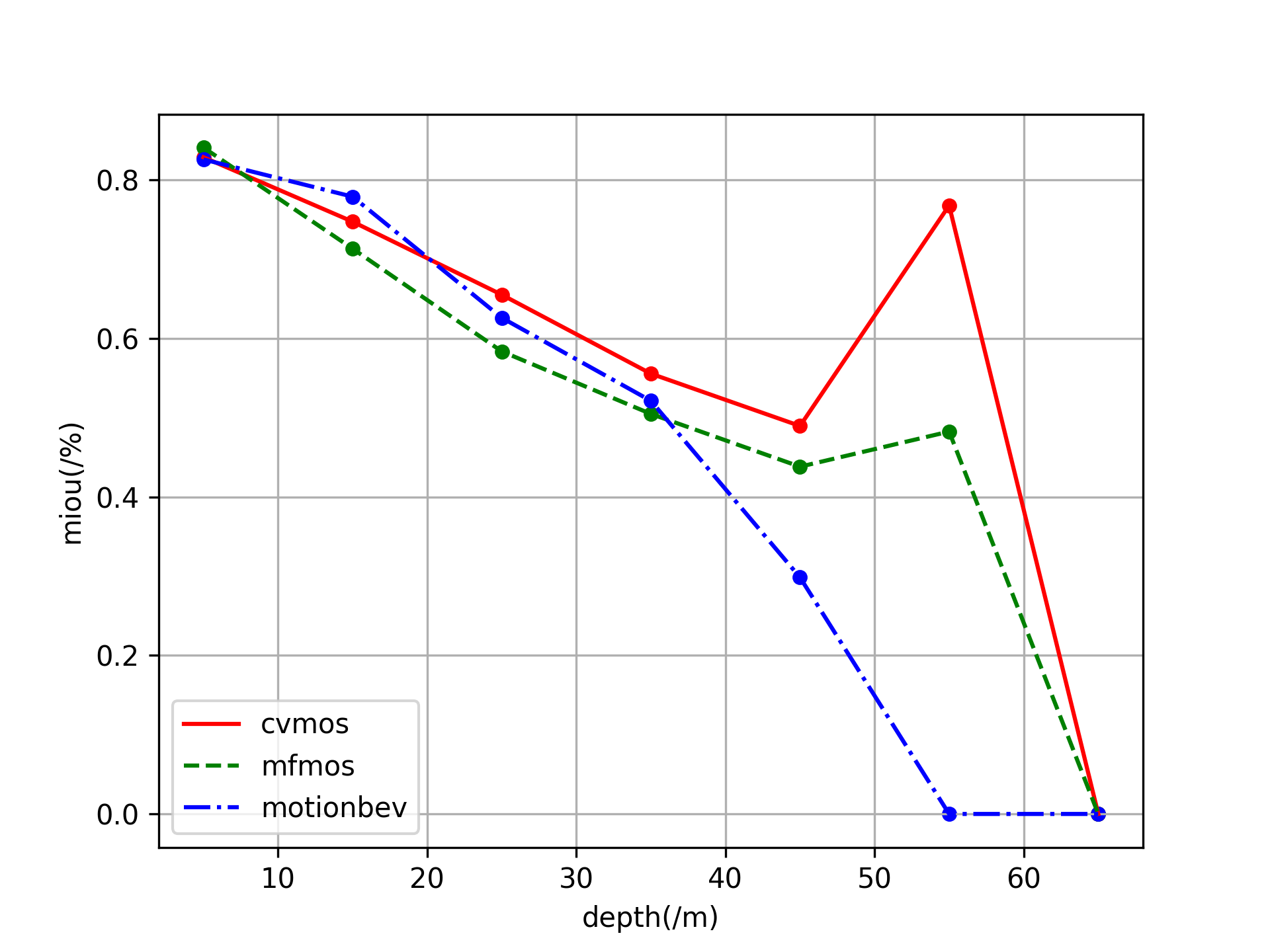}
\hspace{1cm}
% [width=0.8\textwidth,height=0.5\textwidth]
\caption{In the medium to long-distance range of 40-60m, our method performs better than traditional single-view projection methods, and the sparsity of the points has a relatively small impact on our model}
\label{fig: distance}
\end{figure}

For a more intuitive display, we visualized the performance of the model at different distances. As shown in Fig. \ref{fig: distance}, the capability of single-view projection methods to capture moving objects steadily diminishes within the 40-60m range. Conversely, our cross-view projection method consistently maintains high accuracy levels. This suggests that our approach can reduce the impact of the impact of point cloud sparsity.

Furthermore, we report the validation results on the Apollo\cite{lu2019l3} dataset in Fig. \ref{fig: apollo result}. Following previous methods \cite{chen2022automatic, Wang2023InsMOSIM}, we pre-train the CV-MOS model on the SemanticKITTI-MOS dataset and conduct transfer learning and end-to-end fine-tuning for validation experiments on the Apollo dataset. The cross-val setting refers to cross-validation, while the fine-tune setting refers to end-to-end fine-tuning. Compared with other methods based on single-view projection, our CV-MOS performs the best in both settings. Although the Apollo dataset results are not as good as those of 4DMOS\cite{mersch2022receding}, in the test ranking of the large SemanticKITTI-MOS dataset, our method achieves a 14\% higher IoU than 4DMOS\cite{mersch2022receding}, and it also significantly outperforms it in inference speed, which still proves the superiority of our method.

\subsection{Ablation Experiment}
We conducted ablation experiments on the proposed CV-MOS, specifically on the CV motion branch and the SCAM. The results are shown in Tab.\ref{ablation}. Firstly, without any refinement module, our cross-view
motion branch framework proves to be superior. CV-MOS exhibits a significant improvement (+1.1\% IoU) compared to MF-MOS\cite{cheng2024mf}. Additionally, each of the proposed components consistently enhances the performance of our baseline to varying degrees (setting ii). The last row demonstrates that our full CV-MOS achieves the best performance.

\begin{table}
\centering % 表格居中 
\setlength{\tabcolsep}{10pt}
\caption{ABLATION EXPERIMENTS WITH PROPOSED MODULES.}
{\fontsize{10}{14}\selectfont % 将表格中的字体大小设置为较大
\scalebox{1}
{
\begin{tabular}{lllll}
\hline
Methods      & \multicolumn{2}{c}{Component} & IoU(\%) \\ \hline
             & CV            &SCAM           &         \\ \hline
MF-MOS       &               &               & \multicolumn{1}{c}{76.1}    \\
CV-MOS(i)       & \multicolumn{1}{c}{-}             & \multicolumn{1}{c}{-}             & \multicolumn{1}{c}{73.1}    \\ \hline
             & \multicolumn{1}{c}{\textbf{\checkmark}}             & \multicolumn{1}{c}{-}             & \multicolumn{1}{c}{74.2}    \\
CV-MOS(ii)       & \multicolumn{1}{c}{-}              & \multicolumn{1}{c}{\textbf{\checkmark}}             & \multicolumn{1}{c}{76.3}      \\
             & \multicolumn{1}{c}{\textbf{\checkmark}}             & \multicolumn{1}{c}{\textbf{\checkmark}}             & \multicolumn{1}{c}{\textbf{77.5}}    \\ \hline
\end{tabular}
}
}
\label{ablation}
\end{table}

To further validate the efficacy of our modules, we integrated our proposed modules into MotionSeg3D\cite{sun2022efficient}, LMNET\cite{chen2021moving} and MF-MOS\cite{cheng2024mf}, subsequently retraining these models. 

\begin{table}
\centering % 表格居中 
\setlength{\tabcolsep}{5pt}
\caption{THE PROPOSED MODULES PERFORMANCE (\%) ON OTHER METHODS.}
{\fontsize{11}{15}\selectfont % 将表格中的字体大小设置为较大
\scalebox{1}
{
\begin{tabular}{llll}
\hline
Method            & Baseline & w/CV & w/SCAM \\ \hline
LMNet             & \multicolumn{1}{c}{63.82}    & \textbf{+7.02}   & \multicolumn{1}{c}{\textbf{+5.28}}    \\ \hline
Motionseg3D       & \multicolumn{1}{c}{68.07}    & \multicolumn{1}{c}{\textbf{+3.53}} & \multicolumn{1}{c}{\textbf{+0.70}}  \\ \hline
MF-MOS       & \multicolumn{1}{c}{76.12}    & \multicolumn{1}{c}{\textbf{+1.10}} & \multicolumn{1}{c}{\textbf{+0.30}}  \\ \hline
\end{tabular}
}
}
\label{tab-module other method}
\end{table}

\begin{table}[H]
\centering % 表格居中 
\setlength{\tabcolsep}{5pt}
\caption{SCAM Compare With Other Model-V2 INFERENCE TIME (MS) RESULTS.}
{\fontsize{12}{15}\selectfont % 将表格中的字体大小设置为较大
\scalebox{1}
{
\begin{tabular}{llll}
\hline
Method            & Baseline  &  w/SCAM & $\triangle(\downarrow$) \\ \hline
Motionseg3D              & \multicolumn{1}{c}{137}       & \multicolumn{1}{c}{\textbf{131}} & \multicolumn{1}{c}{\textbf{6}}   \\ \hline
MF-MOS       & \multicolumn{1}{c}{137}     & \multicolumn{1}{c}{\textbf{115}} & \multicolumn{1}{c}{\textbf{22}} \\ \hline
\end{tabular}
}
}
\label{tab-scam-infer}
\end{table}

The reported results of adding the CV motion branch to the two networks mentioned above are shown in Tab.\ref{tab-module other method}. It can observe that providing additional BEV motion information to models based on RV projection significantly improves model performance, with a 7.02\% improvement for the LMNet\cite{chen2021moving}. 
Particularly, for the proposed SCAM, we applied it to MotionSeg3D\cite{sun2022efficient} and MF-MOS\cite{cheng2024mf} with the SCAM and tested the inference time on an RTX NVIDIA 3090 machine. The results are shown in Tab.\ref{tab-scam-infer}, $\triangle(\downarrow$) indicates the reduction in inference time (ms) after integrating the SCAM module. As shown in Tab. \ref{tab-module other method} and Tab. \ref{tab-scam-infer}, SCAM not only improves model performance but also accelerates inference speed.

\begin{table*}
    % \centering
    \caption{MOS performance on the SemanticKITTI-MOS validation set for points at different distances. R denotes recall, and P denotes precision.}
    \label{tab: distance}
    \setlength{\tabcolsep}{9pt}  
  \resizebox{1.0\textwidth}{!}{%
    \begin{tabular}{l|ccc|ccc|ccc}
        \hline
        \multirow{2}{*}{\textbf{Method}} & \multicolumn{3}{c|}{\textbf{Close} (\textless20$m$)} & \multicolumn{3}{c|}{\textbf{Medium} (\textgreater=20$m$, \textless 50$m$)} & \multicolumn{3}{c}{\textbf{Far} (\textgreater= 50$m$)}\\  
                                & IoU$_{MOS}$ & R & P       & IoU$_{MOS}$ & R & P       & IoU$_{MOS}$ & R & P \\
        \hline
        \hline
        LMNet~\cite{chen2021moving}&      70.72 & 76.89 & 89.80 &        43.88 & 54.30 & 69.56 &          0.00 & 0.00 & -  \\
        MotionSeg3D~\cite{sun2022efficient}&      71.66 & 79.97 & 87.35 &       52.21 & 59.27 & 81.40 &       4.99 & 4.99 & \textbf{100.00} \\
        MotionBEV~\cite{zhou2023motionbev}&      \textbf{80.85} & 85.40 & \underline{93.81} &      56.35 & 59.89 &  \underline{90.50} &         0.00 & 0.00 & -  \\
        4DMOS~\cite{mersch2022receding}&         78.43 & 82.11 & \textbf{94.59} &         \textbf{68.71} & \textbf{72.62} & \textbf{92.74} &         41.00 & 41.00 & \textbf{100.00} \\
        InsMOS~\cite{Wang2023InsMOSIM}&     75.29 & \textbf{88.78} & 83.21 &       57.67 & 66.81 & 80.84 &         10.88 & 10.89 & \underline{98.63} \\
        MF-MOS~\cite{cheng2024mf}&        79.31 & 84.98 & 92.23 &         54.67 & 64.10 & 78.81 &         \underline{47.97} & \underline{50.08} & 91.94 \\
        \rowcolor{gray!20}
        
        CV-MOS&       \underline{79.87} & \underline{86.46} & 91.30 &         \underline{61.25} & \underline{70.13} & 82.87 &         \textbf{76.78} & \textbf{81.54} & 92.93  \\

        \hline
    \end{tabular}
    }
\end{table*}

\subsection{Qualitative Analysis}
To facilitate a more intuitive comparison between our proposed CV-MOS and other SOTA models based on RV projection \cite{cheng2024mf} and BEV projection \cite{zhou2023motionbev}, we visualized the inference results of the 08 sequence from the SemanticKITTI-MOS dataset, as shown in Fig. \ref{Fig-visualize}. It can be observed that, compared to other SOTA single-view models\cite{zhou2023motionbev, cheng2024mf}, CV-MOS more effectively captures moving objects. This is evident when comparing it with the Ground Truth.

\subsection{Evaluation of Resource Consumption}
We evaluate the inference time (FPS, ms), memory usage (size), and the number of learnable parameters (params) of ours and SOTA methods in Sequence 08 using a single NVIDIA RTX 4090 GPU. We set the batch size to 1 during inference. As shown in Tab.\ref{tab-computation}, compared to the last SOTA model MF-MOS\cite{cheng2024mf}, CV-MOS not only improves accuracy but also accelerates inference speed.

\begin{table}
\centering 
\caption{COMPUTATION RESOURCE COMPARISON.}
\setlength{\tabcolsep}{5pt}
{\fontsize{9}{14}\selectfont % 将表格中的字体大小设置为较大
\begin{tabular}{lcccc}
\toprule %添加表格头部粗线
Method            & FPS & ms  & params(M) & size(MB) \\ \midrule
4DMOS\cite{mersch2022receding}    & 6   & 155  & \textbf{1.84}     &  \textbf{7}    \\
MotionSeg3D-v1\cite{sun2022efficient}    & 18  & 54  & 13.61     &  52    \\
MF-MOS-v1\cite{cheng2024mf}    & \textbf{21}   & \textbf{47}  &  16.70    &  64    \\
CV-MOS-v1       & 19  & 53  & 22.17      & 84     \\ \midrule
MotionSeg3D-v2\cite{sun2022efficient}& 8   & 115  & 35.32     &  132     \\
MF-MOS-v2\cite{cheng2024mf} & 8   & 121 & 35.59     &     132  \\
CV-MOS-v2       & 10   & 102 & 39.07     & 149      \\ \bottomrule %添加表格底部粗线
\end{tabular}
\label{tab-computation}
}
\end{table}

\section{CONCLUSIONS}
This paper proposes a novel and effective online moving object segmentation network based on LiDAR data. By incorporating the concept of cross-view, we expand the original single-view motion branch to cross-view motion branch. Additionally, we introduce the spatial and channel attention module to mitigate information loss from coarse to fine levels. Extensive experiments demonstrate that 1) our proposed CV-MOS achieves the highest accuracy on both validation and test sets of the SemanticKITTI-MOS Dataset. 2) The proposed model exhibits superior performance and generalization ability, making it suitable for use with other projection-based methods.

\bibliographystyle{IEEEtran}
\bibliography{referrence}
	% Author1
\newpage

\begin{IEEEbiography}[
	{\includegraphics[width=0.81in,height=1in,clip,keepaspectratio]{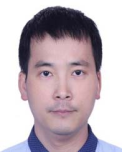}}]{Xiaoyu Tang}
        \scriptsize
	(Member, IEEE) received the B.S. degree from South China Normal University in 2003 and the M.S. degree from Sun Yat-sen University in 2011. He is currently pursuing the Ph.D. degree with South China Normal University. He is working with the School of Physics, South China Normal University, where he engaged in information system development. His research interests include machine vision, intelligent control, and the Internet of Things. He is a member of the IEEE ICICSP Technical Committee.
    
\end{IEEEbiography}

\vspace{-1.3cm}

% Author2
\begin{IEEEbiography}[
	{\includegraphics[width=0.81in,height=1in,clip,keepaspectratio]{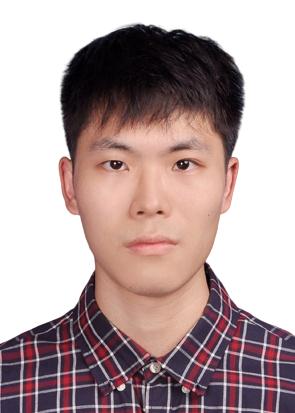}}]{Zeyu Chen}
    \scriptsize
	received the B.Eng. degree from the School of Physics and Telecommunication Engineering, South China Normal University, in 2022, where he is currently pursuing the M.S. degree with the Department of Electronics and Information Engineering. His research is computer vision and deeplearning.
\end{IEEEbiography}

\vspace{-1.3cm}

% Author3
\begin{IEEEbiography}[
    {\includegraphics[width=0.81in,height=1in,clip,keepaspectratio]{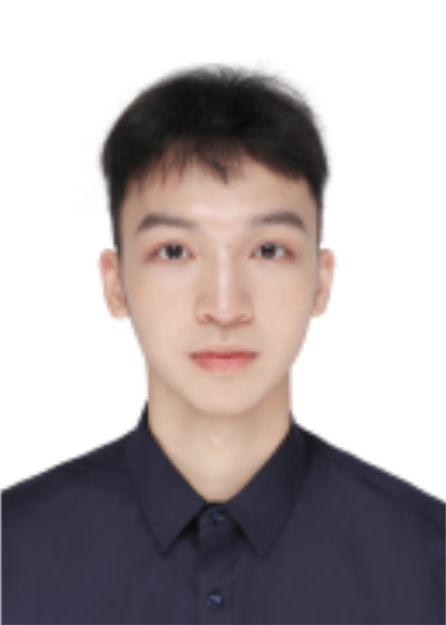}}]{Jintao Cheng}
    \scriptsize
    received his bachelor's degree from the School of Physics and Telecommunications Engineering, South China Normal University in 2021. His research is computer vision, SLAM and deeplearning.
\end{IEEEbiography}

\vspace{-1.3cm}

% Author4
\begin{IEEEbiography}[
	{\includegraphics[width=0.81in,height=1in,clip,keepaspectratio]{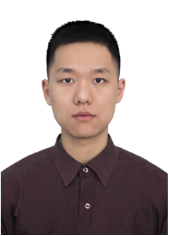}}]{Xieyuanli Chen}
 \scriptsize
Xieyuanli Chen received his Ph.D. degree in
Robotics at Photogrammetry and Robotics Laboratory, University of Bonn. He is now also a member of the Technical Committee of RoboCup Rescue Robot League (RRL). He received his Master degree in Robotics in 2017 at the National University of Defense Technology, China. During that time, he was a member of the Organizing Committee of RoboCup Rescue Robot League. He received his Bachelor degree in Electrical Engineering and Automation in 2015 at Hunan University, China.
\end{IEEEbiography}

\vspace{-1.3cm}

    % Author5
\begin{IEEEbiography}[
	{\includegraphics[width=0.81in,height=1in,clip]{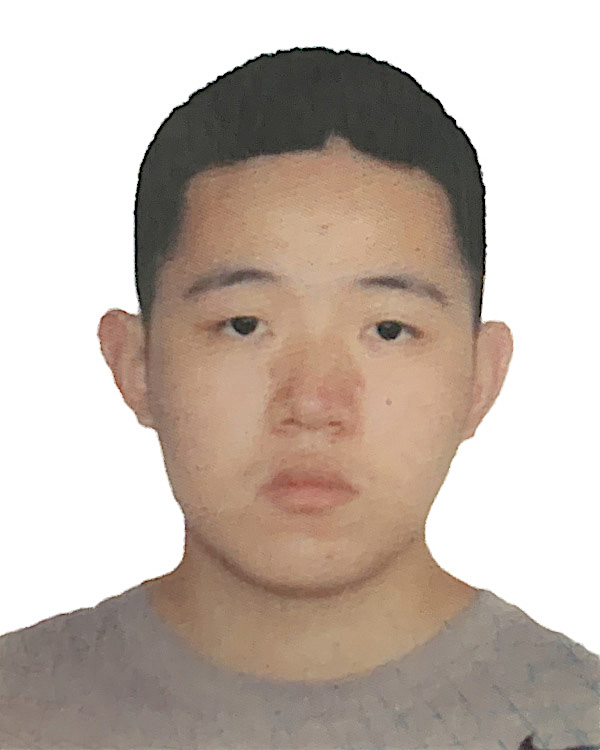}}]{Jin Wu}
 \scriptsize
(Member, IEEE) was born in Zhenjiang, China, in 1994. He received a B.S. degree from the University of Electronic Science and Technology of China, Chengdu, China. From 2013 to 2014, He was a visiting student with Groep T, Katholieke Universiteit Leuven (KU Leuven). He is currently pursuing a Ph.D. degree in the Department of Electronic and Computer Engineering, Hong Kong University of Science and Technology (HKUST), Hong Kong. He has co-authored over 120 technical papers in representative journals and conference proceedings. He was awarded the outstanding reviewer of \textsc{IEEE Transactions on Instrumentation and Measurement} in 2021. He is now a Review Editor of Frontiers in Aerospace Engineering and an invited guest editor for 5 special issues of MDPI. He is also in the IEEE Consumer Technology Society (CTSoc), as a committee member and publication liaison. He was a committee member for the IEEE CoDIT conference in 2019, a special section chair for the IEEE ICGNC conference in 2021, a special session chair for the 2023 IEEE International Conference on Intelligent Transportation Systems (ITSC), a Track Chair for the 2024 IEEE International Conference on Consumer Electronics (ICCE) and a Chair for the 2024 IEEE CTSoc Gaming, Entertainment and Media (GEM) conference. He was selected as the World’s Top 2$\%$ Scientist by Stanford University and Elsevier, in the 2020, 2021 and 2022 year round.
\end{IEEEbiography}

\vspace{-1.3cm}

    % Author6
\begin{IEEEbiography}[
	{\includegraphics[width=0.81in,height=1in,clip]{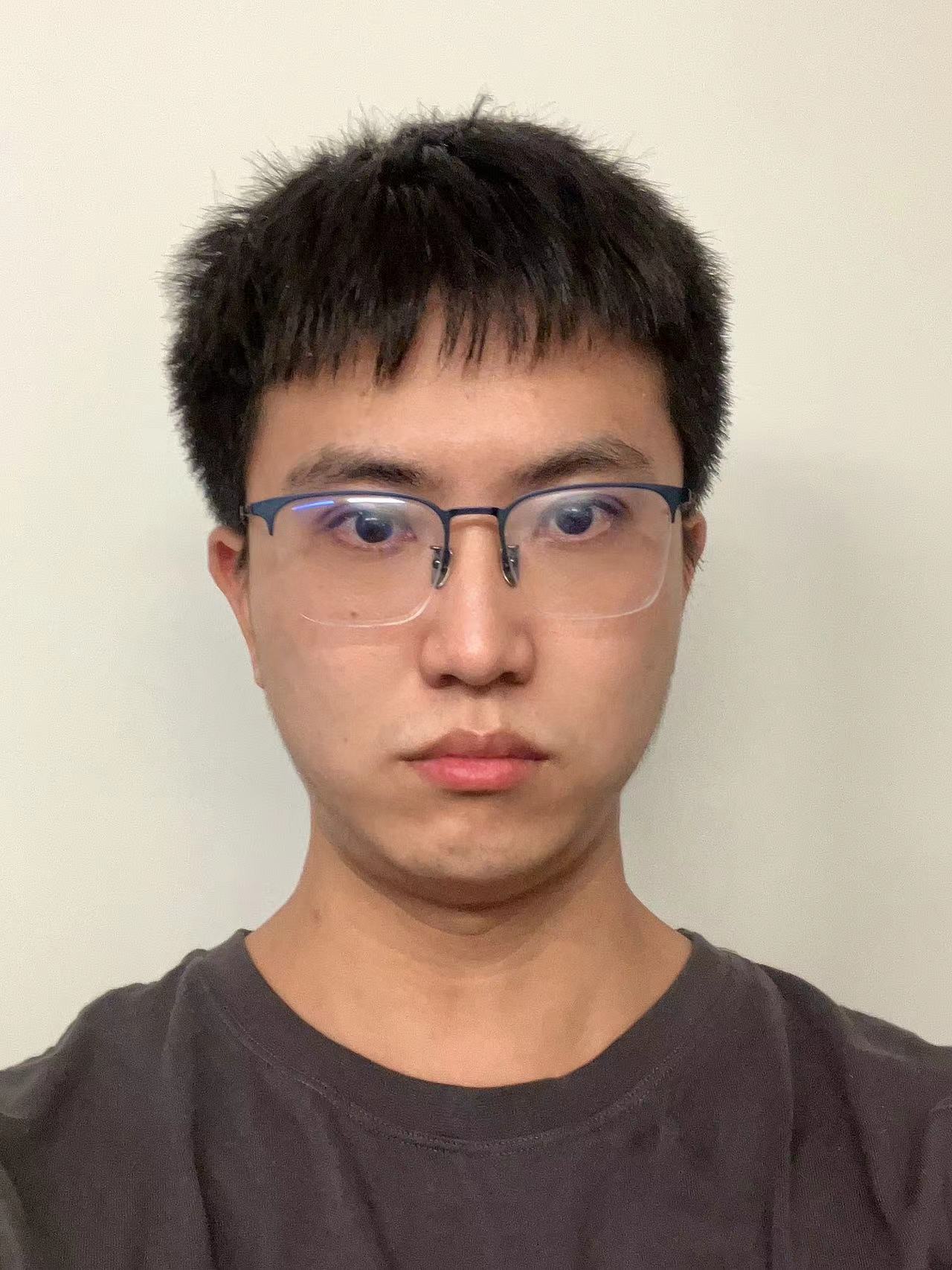}}]{Bohuan Xue}
    \scriptsize
Bohuan Xue (Graduate Student Member, IEEE) received the B.Eng. degree in computer science and technology from College of Mobile Telecommunications, Chongqing University of Posts and and Telecom, Chongqing, China, in 2018. He is currently working toward the Ph.D. degree in electrical engineering with the Department of Computer Science and Engineering, Hong Kong University of Science and Technology, HKSAR, China. His research interests include SLAM, computer vision, and 3D reconstruction.
\end{IEEEbiography}

\end{document}